\documentclass{ifacconf}
\usepackage{amsmath,amssymb,amsfonts,color,soul}
\usepackage{textcomp, natbib, comment}
\usepackage{xcolor}
\usepackage{graphicx}      % include this line if your document contains figures
\usepackage{natbib}        % required for bibliography
\begin{document}
\begin{frontmatter}

\title{Collision Avoidance Using Spherical Harmonics\thanksref{footnoteinfo}} 

\thanks[footnoteinfo]{This research has been supported in part by NSF award ECCS-1924790.}

\author[UT]{Steven D. Patrick}
\author[UT]{Efstathios Bakolas}

\address[UT]{The University of Texas at Austin, Austin, Texas 78712-1221 (spatric5@utexas.edu, bakolas@austin.utexas.edu).}

\begin{abstract}                % Abstract of not more than 250 words.
In this paper, we propose a novel optimization-based trajectory planner that utilizes spherical harmonics to estimate the collision-free solution space around an agent. The space is estimated using a constrained over-determined least-squares estimator to determine the parameters that define a spherical harmonic approximation at a given time step. Since spherical harmonics produce star-convex shapes, the planner can consider all paths that are in line-of-sight for the agent within a given radius. This contrasts with other state-of-the-art planners that generate trajectories by estimating obstacle boundaries with rough approximations and using heuristic rules to prune a solution space into one that can be easily explored. Those methods cause the trajectory planner to be overly conservative in environments where an agent must get close to obstacles to accomplish a goal. Our method is shown to perform on-par with other path planners and surpass these planners in certain environments. It generates feasible trajectories while still running in real-time and guaranteeing safety when a valid solution exists.
\end{abstract}

\begin{keyword}
Collision avoidance, path planning, spherical harmonics
\end{keyword}

\end{frontmatter}
%===============================================================================

\section{Introduction} \label{Intro}
\begin{figure}
    \centering
    \includegraphics[scale=.25]{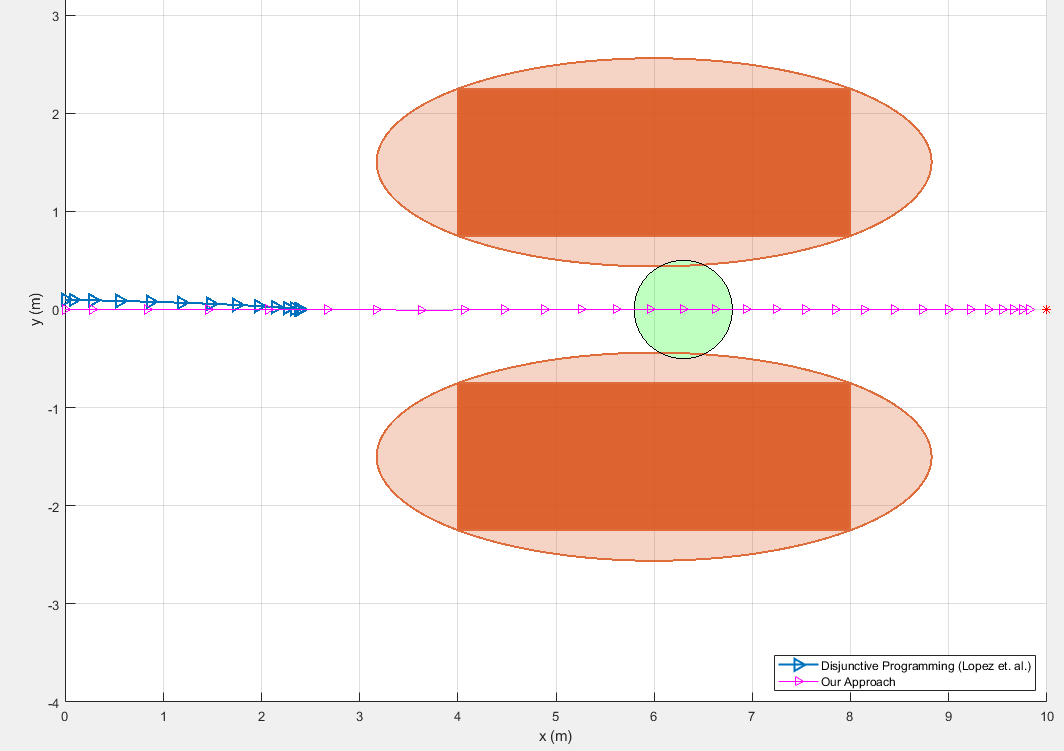}
    \caption{Our approach compared to \cite{castillo2020real} for an agent trying to pass through a narrow corridor. Orange are the rectangular obstacles. Light orange are their bounding ellipsoids. Blue is their path, and magenta is ours. Green is the agent at a given time step. The agent clearly overlaps the light orange ellipses, but does not touch dark orange boxes.}
    \label{fig:Pretty_Fig}
\end{figure}

In this paper, we introduce a path planner that is capable of finding feasible trajectories in an environment with no apriori information about the agent's surroundings. 
The planner generates trajectories by gathering 3D point cloud data from the environment to approximate a collision-free space using spherical harmonics (SH). 
The collision-free space determines the domain of feasible trajectories. 
Within the domain, an optimization-based trajectory planner determines the optimal trajectory that minimizes a cost function while satisfying other practical constraints such as velocity thresholds or control input limits.
The process of measurement to feasible trajectory happens in real-time and runs in a continuous loop until the agent accomplishes its goal.
Using SH to estimate the collision-free space allows our path planner to generate less conservative trajectories than other methods that approximate individual obstacles with simplified geometries such as ellipsoids or bounding boxes. An example can be seen in Fig. \ref{fig:Pretty_Fig}.

\textit{Literature review:} Path and motion planning problems have received considerable attention in the literature. Methods that rely on machine learning tools are considered among the most prevalent ones in the field at present. 
There have been attempts to create a path planner using vision data and convolutional neural networks to produce valid trajectories \cite{chakravarty2017cnn} or reinforcement learning to determine effective maneuvering policies after numerous training iterations \cite{everett2021collision}. 
However, these approaches rely heavily on the training data they are exposed to. 
If a completely new environment is introduced to the system, it is hard to know if the system will be able to cope with the increased uncertainty \cite{eykholt2018robust}. 
Because of ambiguity in why decisions are made in machine learning approaches, hard programmed methodologies are used where certain safety standards are required. 
Rapidly-exploring Random Tree (RRT) is another data-driven approach where a trajectory planner samples trajectories in waves with each wave of trajectory sampling being a variation of the previous wave \cite{lavalle1998RRT}. RRT requires full knowledge of the surrounding area. Additionally, RRT requires the obstacles to be stationary in order to plan a full trajectory. If anything changes in the environment, then a new trajectory must be planned. Due to the sampling nature of this planner, it is difficult to implement in real-time. However, there are approaches that have tried to overcome the limitations of RRT such as \cite{kuffner2000rrt_star} with RRT* or \cite{naderi2015rt} with RT-RRT*. The former is still not considered real-time, but it can handle dynamic obstacles. The latter is real-time and can handle dynamic obstacles, but it requires large amounts of memory and relatively simple environments. Both approaches still require near-perfect knowledge of the environment which is something that cannot always be guaranteed in the real-world.
Another popular method for collision avoidance is artificial potential fields (APF) \cite{khansari2012dynamical} \cite{marchidan2020collision}. 
APF approaches rely on an artificial flow running through a given environment. 
Obstacles interrupt the artificial flow and cause free particles around them to move away. 
Therefore, an agent’s position can be treated as a free-floating particle in the environment, and it should move within the flow of the system. 
Because of the path being passive to the environment, the agent has the potential of getting stuck in a stable-point or vortex that is not the goal state \cite{janabi1993integration}.
An additional drawback of the approach is the artificial flow must be discretized for real-world applications.
%the action space (set of actions available to the agent that determine its direction of motion at each time). 
If the discretization is too coarse, the planned path could run into an obstacle.

Optimization-based approaches in path / motion planning do not suffer from the aforementioned drawbacks and have shown promising results in recent years. 
These methods involve a path planner minimizing a cost function associated with completing a task subject to constraints and other mission specifications. 
Unlike machine learning, optimization does not need to be retrained for different environments, and it is easier to diagnose for its decision making process.
Furthermore, the optimization-based approach can provide guarantees in terms of performance and maintaining safety. 
However, the resulting trajectories through optimization are highly dependent on the cost function and constraints, so the formulation of these statements is critical to the success or failure of the path planner.
As in \cite{lin2020robust}, researchers have tried to create a search space of valid trajectories by intersecting half-spaces generated by obstacles. 
Another approach attempted to use disjunctive programming for a collision avoidance constraint \cite{castillo2020real}. 
Both of these approaches rely on ellipsoidal approximations of obstacles which produce undesirable results in scenarios where an agent must get close to an obstacle or even interact with its environment. 
Our motivation for this paper is to address the issues associated with coarse ellipsoidal approximations.  
With SH estimation of the collision-free space, our optimization-based path planner can produce less conservative trajectories than other state-of-the-art planners.

\textit{Outline of the paper:} %In Section \ref{Intro}, some related works are discussed in addition to why our proposal is necessary. 
In Section \ref{Prob_Stat}, we formulate our optimization-based collision avoidance problem for local motion planning and compare our approach with other state-of-the-art methods in the field. 
The fundamental concepts of our approach are explained in Section \ref{Bkgrnd}. 
In Section \ref{Contrib}, we showcase our contribution and explain in detail how we implement the approach for a general system. 
Section \ref{Results} demonstrates the efficacy of our approach in some defining examples via numerical simulations and also presents comparisons of our approach to other methods. Finally, Section~\ref{s:conclusion} presents concluding remarks and directions for future research.

\section{Problem Statement}\label{Prob_Stat}

\subsection{Optimization for Trajectory Planning}\label{s:Opt_Bckgnd}
We assume that the motion of the agent is described by the following discrete-time state space model:
\begin{align}
    x_{t+1}&=f(x_t,u_t)+w_t,
\end{align}
where $x_t\in \mathbb{R}^p$ denotes the state of the robot agent at time $t$, $u_t \in \mathbb{R}^m$ is the control input at time $t$, the function $f$ represents the dynamics (vector field) of the agent, and $w_t \in \mathbb{R}^p$ is the process noise that acts upon the agent at time $t$.

The main goal of an optimization-based trajectory planner is to minimize a cost function while satisfying various equality and inequality constraints.
The cost function can be the amount fuel used, time traveled, and/or distance from goal. 
Examples of inequality constraints include maximum accelerations, maximum velocities, and valid orientations. 
A popular methodology for optimization-based trajectory planning is model predictive control.
It works by inducing a constraint on the planner to abide by the dynamics of the system.
The benefit of model predictive control is the planner can determine multiple inputs at subsequent time steps within a given time horizon in a single optimization problem. 
The constrained optimization can be formulated as follows: 

\begin{subequations}
\begin{align}
    \min_{u_0,\dots,u_{N-1}} &J(u_0,\dots,u_{N-1},x_0,\dots,x_N) \\ \label{MPC_ego}
    x_{t+1} &= f(x_t,u_t)+w_t \\ \label{Gen_CA_Cnstr}
     g_{lb} &\leq g(x_t) \leq g_{ub} && \forall t \in [0,N]\\ \label{State_Cnstr}
    x_{i} &\in \mathbf{X}, u_{i-1} \in \mathbf{U} && \forall t \in [1,N]
\end{align}
\end{subequations}

The cost function, $J$, takes in the input arguments of the state and control input at discrete time steps along the (discrete) time intervals $[0,N]$ and $[0,N-1]$, respectively (here, by $[0,N]$ we denote the discrete set $\{0,\dots,N \})$. 
Equation (\ref{MPC_ego}) is the constraint for the previously mentioned model predictive control formulation for path planning. 
The next constraint, Equation (\ref{Gen_CA_Cnstr}), is defined as the collision avoidance constraint. 
It states that the state at time $t$ must satisfy the collision avoidance constraint. 
Equation (\ref{State_Cnstr}) states that the state and control input must take values in sets $\mathbf{X}$ and $\mathbf{U}$ respectively. 
This is equivalent to, for instance, enforcing min-max acceleration and velocity values as well as defining invalid orientations of an agent.

\subsection{Obstacle Representation} \label{s:Obs_Rep}
\begin{figure}[htbp]
\centerline{\includegraphics[scale=.32]{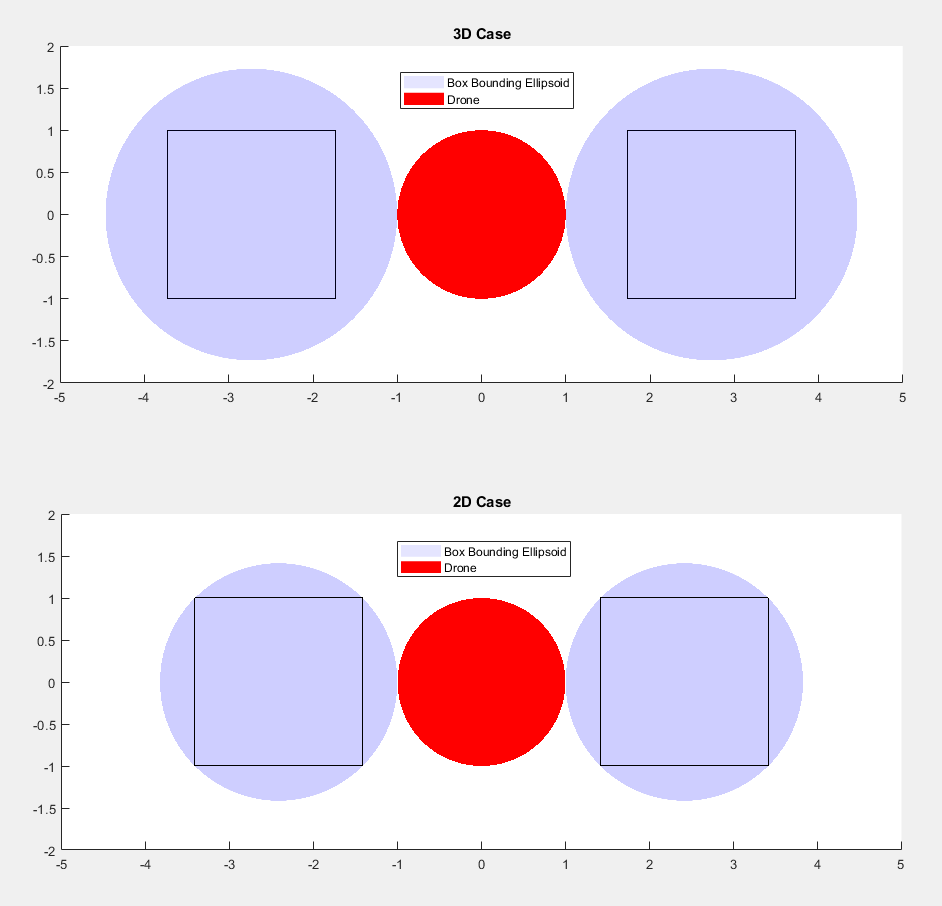}}
\caption{Demonstration of the amount of space required for an agent to pass through a narrow passage. The top sub-figure is the 3D case and the bottom sub-figure is the 2D case for a bounding ellipsoid. The agent is in red with radius 1, the obstacles true shape is the black outline, and the obstacles bounding ellipsoids are in blue. }
\label{Neccesary_Gap}
\end{figure}

As mentioned in Section \ref{s:Opt_Bckgnd}, optimization-based trajectory planning requires the inclusion of a collision avoidance constraint to avoid obstacles within a given environment. 
If an obstacle is represented as a convex set, its complement space, or the solution space for trajectory planning, is non-convex. 
To make the solution space convex, a common approach is to divide the space into half spaces. 
These half spaces are defined by tangent planes along an obstacle's boundary.
The intersection of multiple half spaces can be used to define a convex region of valid solutions for the trajectory planner to choose from. 
To produce these half spaces, obstacles are typically approximated as bounding boxes or ellipsoids. 
The approximation leads to the collision avoidance constraint padding around the obstacles. 
The padding causes the path planner to produce trajectories that are unnecessarily far away from obstacles.  
A defining example where padding can produce an undesirable trajectory is a small hallway like opening as seen in Fig. \ref{Neccesary_Gap} where the agent must pass through the narrow gap. 
Even though the agent can make it between the obstacles, the optimizer will produce a path that either goes around the hallway or freezes at the entry of the hallway.

To further exemplify the problem, an agent with radius $r_a$ to pass between two square boxes with side width $w_b$ requires a gap equal to $2r_a+w_b(\sqrt{2}-1)$ for the 2D case and $2r_a+w_b(\sqrt{3}-1)$ for the 3D case. 
Notice that the clearance depends on the width of the obstacle. Therefore, a wider obstacle requires a greater gap for an agent to pass through it. 
Ideally, the gap an agent can pass through should only depend on the agent's dimensions.

\subsection{Major Contribution}
Because of the limitations mentioned in Section \ref{s:Obs_Rep}, we propose a new collision avoidance constraint for an optimization-based approach for trajectory planning. 
Our collision avoidance constraint requires the agent to stay inside of a collision-free space at all given time steps in the planning horizon. 
The collision-free space is well defined by using a least squares estimator to estimate the parameters of a space defined by the summation of SH basis functions. 
Using our SH approximation of the collision-free space for a collision avoidance constraint allows the trajectory planner to be less conservative than other approaches around obstacles, but still maintains degrees of safety, real-time capabilities, and diagnosable actions.

\section{Background}\label{Bkgrnd}

\subsection{Spherical Harmonics}
For 1D signal estimation, the Fourier transform is a fundamental tool for reconstructing a measured periodic signal. 
A Fourier transform comprises of a weighted sum of sine and cosine basis functions of different frequencies to approximate the measured signal. 
Spherical Harmonic (SH) estimation is the natural extension of this idea to three dimensions. 
The three dimensional basis functions are the eigenfunctions of the angular portion of the Laplacian \cite{kurz2020three}. 
In order to get the angular portion of the Laplacian, a conversion from Cartesian to spherical coordinates and its inverse are required. 
The basic transformation from spherical coordinates to Cartesian is described by the following equations:
\begin{align}\label{RPT2X}
    x &= r \sin(\theta)\cos(\phi), \\\label{RPT2Y}
    y &= r \sin(\theta)\sin(\phi), \\\label{RPT2Z}
    z &= r \cos(\theta),
\end{align}
where $\phi \in [0,2\pi], \theta \in [0,\pi]$, and $r \geq 0$.
Using polar coordinates allows any 3D shape to be represented as a periodic function $r(\theta,\phi)$ with period $\pi$ and $2\pi$ for $\theta$ and $\phi$ respectively.
The approximation of the 3D space is the sum of the basis functions, $Y^m_\ell(\theta,\phi)$, multiplied by their respective weights:
\begin{equation}\label{R_SH}
    r_{SH}(\theta,\phi) = \sum_{l=0}^\infty \sum_{m=-\ell}^\ell \alpha_\ell^m Y_\ell^m(\theta,\phi).
\end{equation}

The summation index $\ell$ corresponds to the order of the harmonic and the index $m$ goes through the individual harmonics inside an order.
These basis functions output a radius value for a given $\theta$ and $\phi$.
It can be seen that the order of estimation and the parameter space of the spherical harmonic is quadratic: $\sum_{i=1}^{L} 2i-1 = L^2$ where $L$ is the highest order of estimation. 
This is in contrast to a 1D Fourier series which has a linear relationship: $\sum_{i=1}^{L} 2 = 2L$.

\begin{figure}
    \centering
    \includegraphics[scale=.25]{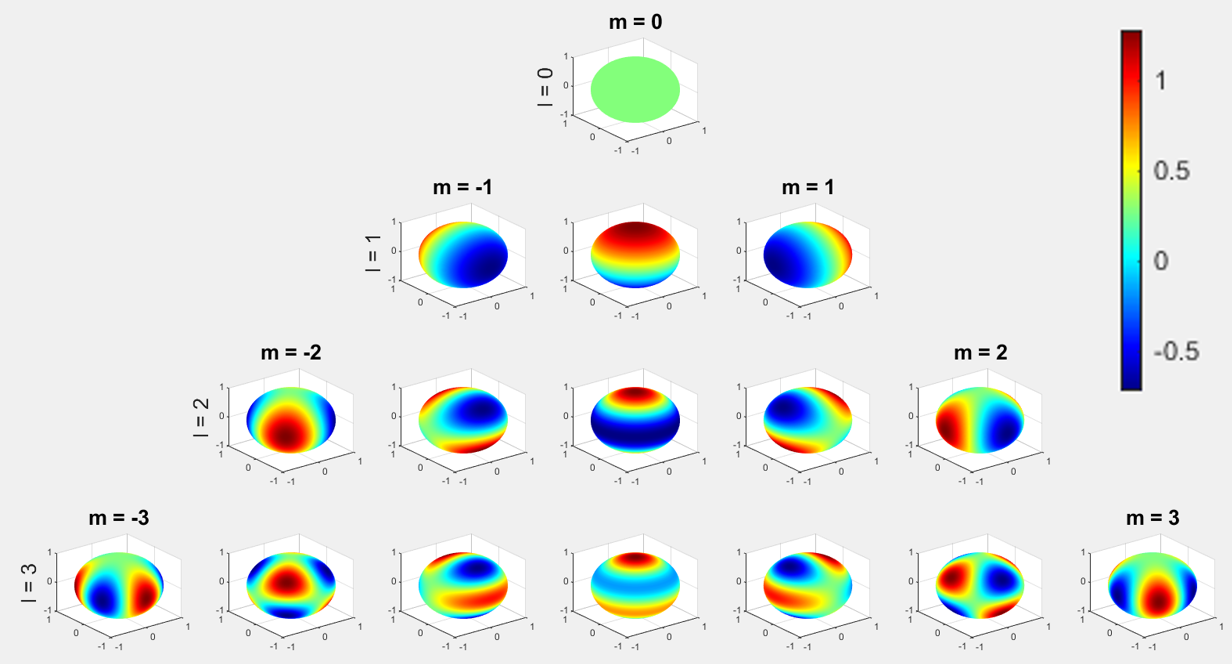}
    \caption{Spherical Harmonic basis functions for orders 0 through 3. The colors indicate the value of the radius at that point.}
    \label{fig:SH_Weights}
\end{figure}

In addition, the basis functions are defined as follows:
\begin{equation}
    Y_\ell^m(\theta,\phi) = \sqrt{\frac{(2\ell+1)(l-m)!}{4\pi(\ell+m)!}} P_\ell^m(\cos(\theta)) \mathrm{e}^{im\phi},
\end{equation}
where $P^m_l(x)$ are the Legendre Polynomials:

\begin{align}
    P_\ell^m(x) &= \frac{(-1)^m}{\ell!2^\ell}(1+x^2)^{\frac{m}{2}} \frac{d^{\ell+m}}{dx^{\ell+m}} (x^2-1)^\ell.
\end{align}
For this application, only the real part of the basis functions are used. Examples of some SH basis functions can be seen in Fig. \ref{fig:SH_Weights}. 

There are several ways to estimate the weights for Equation (\ref{R_SH}) such as least squares \cite{shen2009modeling}, Unscented Kalman Filter \cite{kurz2020three}, or support vector machine \cite{medyukhina2020dynamic}. Herein, we use a variation on a least squares approach. The methodology is discussed in Section \ref{s:CFS_Est}.

\section{Free Space Constraint}\label{Contrib}

\subsection{Data Retrieval and Pre-Processing}\label{PreProcessing}
\begin{figure}[htbp]
\centerline{\includegraphics[scale=.5]{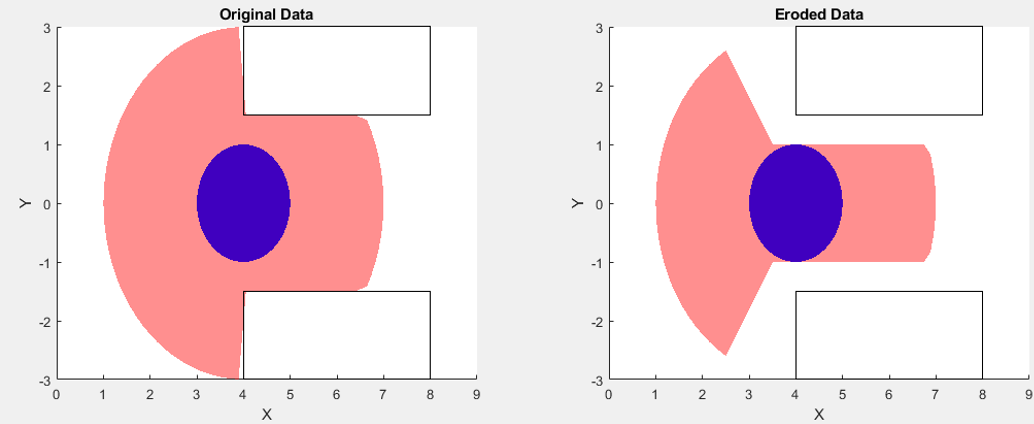}}
\caption{ Example of raw data being concatenated into a meaningful radius. Then offsetting the result by the agent's radius.}
\label{Data_PreProcessing}
\end{figure}

Conventional 3D point cloud generators like LIDAR or UV-difference map do not produce convenient data to work with for SH estimation. 
To account for this, the system converts the collected points and erodes the point cloud into a sphere of meaningful radius. 
This means that if a generated point is outside a given radius of concern, it is interpolated inwards towards the agent's inertial body frame. 
An ideal radius to erode points to is equal to the maximum allowed distance the agent can travel within the given time horizon plus the radius of the agent. 
After this consolidation of points into a region of interest, the points are then eroded further by the agents radius. 
The resulting point cloud is the agent's collision free space and on the interior of the space would not cause a collision with any object. 
An example of the pre-processing steps can be seen in Fig.\ref{Data_PreProcessing}.

\subsection{Collision-Free Space Estimation}\label{s:CFS_Est}
In this paper, we use least squares estimator to get the weights similar to \cite{kurz2020three}. 
The key difference is we add constraints to our least squares estimator:
\begin{subequations}\label{LS_Weights}
    \begin{align}
        \min_{x} &\|Ax-b\|^2_2, \\ \label{eq:Hard_points}
        \textbf{subject to  } & 0 \leq Cx\leq d, \\ \label{eq:Soft_points}
        &0 \leq Ax \leq b, \\ \label{eq:State_cnstr}
        & h_{lb} \leq x \leq h_{ub},
    \end{align}
\end{subequations}
The cost function is the squared $\ell_2$-norm of the error of an affine transformation.
The error, in this case, is the difference of radii between a sphere and the resulting resulting SH estimation of the collision-free space.
The $b$ matrix is a column vector with length equal to the number of points, $n$, sampled along a spherical boundary. 
Its values are a repeated constant of the desired radius of the free-space.
In particular, we take the radius to be equal to the maximum distance the agent could travel within a given time horizon. In particular,
\begin{align}
    b &= r\cdot \begin{bmatrix}
    1 & 1 &\cdots & 1
    \end{bmatrix}^{\mathrm{T}} \in \mathbb{R}^{n\times 1}\\
    x &= \begin{bmatrix}\alpha_0^0 & \alpha_1^{-1} & \cdots & \alpha_{L}^{L}\end{bmatrix}^{\mathrm{T}} \in \mathbb{R}^{L^2 \times 1}
\end{align}
The entries of $A$ correspond to the values of the $Y^m_l$ functions at given points along a surface of a desired spherical free-space.
\begin{equation}
    \textstyle A = \begin{bmatrix}
     \scriptstyle Y_0^0(\theta_1,\phi_1) & \scriptstyle Y_1^{-1}(\theta_1,\phi_1) & \cdots & \scriptstyle Y_L^{L}(\theta_1,\phi_1) \\
    \scriptstyle Y_0^0(\theta_2,\phi_2) & \scriptstyle Y_1^{-1}(\theta_2,\phi_2) & \cdots &  \scriptstyle Y_L^L(\theta_2,\phi_2) \\
    \vdots & \vdots & \ddots & \vdots \\
    \scriptstyle Y_0^0(\theta_n,\phi_n) & \scriptstyle Y_1^{-1}(\theta_n,\phi_n) & \cdots & \scriptstyle Y_L^L(\theta_n,\phi_n)
    \end{bmatrix} \in \mathbb{R}^{n\times L^2}
\end{equation}
The column index, $j$, is determined by $j=l^2+l+m$.
The column matrix $x$ are the weights for a given spherical harmonic.
The minimum of this function without any constraints will happen when the first entry in $x$ is the radius of the desired sphere divided by $Y_0^0 = \sqrt{\frac{1}{4\pi}}$.
The $C \in \mathbb{R}^{m\times L^2}$ and $d \in \mathbb{R}^m$ matrices are defined similarly to $A$ and $b$, but they correspond to points from the pre-processed data discussed in Section \ref{PreProcessing} where $m$ corresponds to the number of measured points.

\begin{figure}[htbp]
\centerline{\includegraphics[scale=.3]{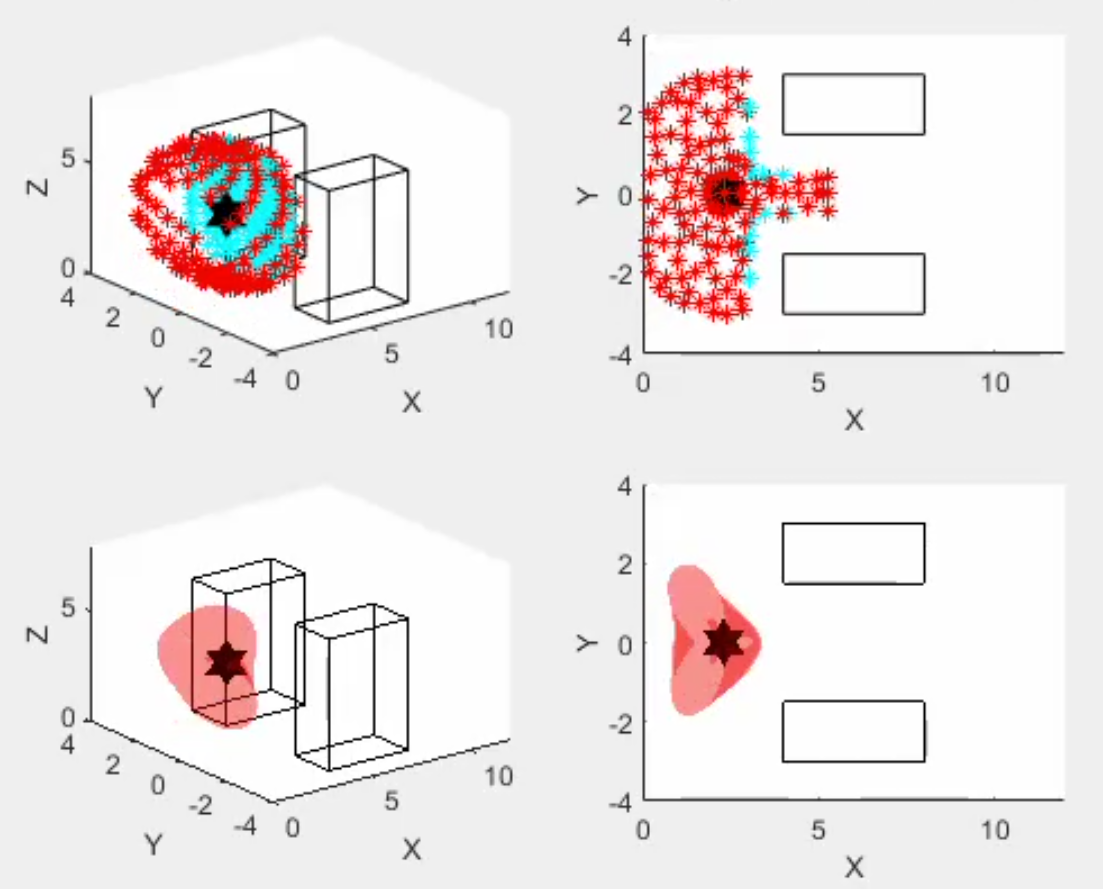}}
\caption{The top row is an example of the points used in least squares estimator. The agent's location is the black star and the black outlines are the obstacles. Cyan points are the hard constraint points that are incorporated into matrices $C$ and $d$. The red points are the points that encourage a spherical shape for matrices $A$ and $b$. The bottom shows the resulting spherical harmonic with the given points. The left column is a 3D view and the right column is a view from above. }
\label{LS_Example}
\end{figure}

Equation (\ref{eq:Hard_points}) comes from the measured points from a given measurement unit.
Equation (\ref{eq:Soft_points}) bounds the overall shape inside a sphere.
These are the same points that are used in the cost function, but the Equation (\ref{eq:Soft_points}) guarantees the solution does not produce extreme spikes.
It is also important to have a lower bound for Equation (\ref{eq:Hard_points}) and (\ref{eq:Soft_points}) to be a non-negative number, so the radius of the estimated 3D shape at the given values $\phi_i$ and $\theta_i$ will be positive.
The element wise inequality in Equation (\ref{eq:State_cnstr}) guarantees reasonable values for the weights to achieve a smooth SH estimation. 
It can be empirically determined to have the upper bound for any weight in the summation to be four times the maximum radius the agent can travel within the given time horizon.
An illustrative example of these points can be seen in Fig. \ref{LS_Example}.

\begin{figure*}
    \centering
    \includegraphics[width=.9\textwidth]{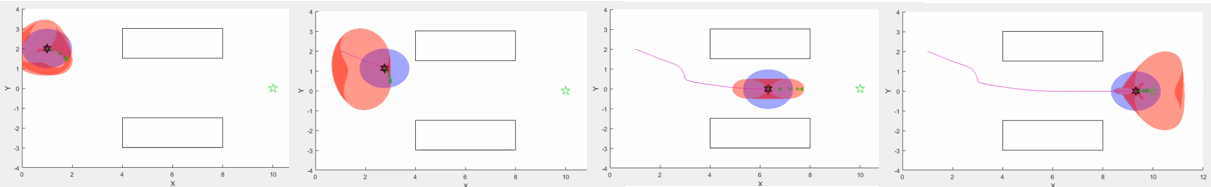}
    \caption{Snapshot of agent throughout time. Red is the spherical harmonic approximation of the collision-free space. Blue is the bounding region of the agent. The green star is the goal. The green path is the planned path over the time horizon. The white boxes are stationary obstacles. The magenta path is the final path taken.}
    \label{fig:Path_Over_Time}
\end{figure*}

\subsection{System Setup}
The pipeline starts by retrieving the 3D point cloud data. From there, the pre-processing occurs as described in Section \ref{PreProcessing}. 
The points are used to create a spherical harmonic estimation of the free space at the given time step. 
The collision-free space estimation is then used for the collision avoidance constraint. 
The position of the agent must be inside of the spherical harmonic for all the time steps. 
In addition, the trajectory must satisfy other necessary constraints such as valid input values and ones required for model predictive control implementation while minimizing the distance from the goal position and energy used to get there. 
Next, we formulate the optimization problem:
\begin{subequations}\label{LS_Opt}
\begin{align}
    \min_{u_0,\cdots,u_{N-1}} &J(u_0,\dots,u_{N-1},x_0,\dots,x_N) \\
    x_{t+1} &= f(x_t,u_t)+w_t \\
    r_{SH}(x_t)&-r_{x_t} \geq 0 \\ \label{CA_Constr}
    x_i &\in \mathbf{X},~ u_{i-1} \in \mathbf{U} ~~~~~~~~~ \forall t \in [1,N].
\end{align}
\end{subequations}

Since the primary focus of this paper is ensuring that the collision avoidance constraint is satisfied at all times, a simple cost function of the distance from the goal and total energy is used:
\begin{equation}
    J = \sum_{t=1}^N (\|x_t-x_g\|^2_P+\|u_{t-1}\|^2_Q),
\end{equation}
where $x_g$ is the goal position, and the norms are weighted using positive definite matrices $P$ and $Q$.

\section{Case Study: Aerial Vehicle}\label{Results}

\subsection{Case Study Setup}
For this example, a drone tries to pass between two obstacles that are relatively close together. 
Our goal is to showcase the agent choosing the shortest path by going through the narrow passage rather than around the two rectangular obstacles or freezing. 
To demonstrate the obstacle avoidance capabilities, we placed the agent's starting position at the same y-location as one of the obstacles. 
Additionally, the obstacles are placed so the agent has less than its radius clearance between the two obstacles.

For the dynamics of the system, we use drone dynamics for the agent as in \cite{castillo2020real} and \cite{lin2020robust}. 
The state of the drone is its position in 3D space $p$, heading $\psi$, linear velocity $v$, and angular velocity $\Dot{\psi}$. All obstacles are treated as stationary, so the state of the obstacle is its position with velocity equal to zero.
The equations of motion for the system are:
\begin{subequations}
    \begin{align}\label{Pos_Der}
        \Dot{p} &= R(\phi)v,\\
        \Dot{v} &= \frac{-v+ku}{\tau},\\ \label{AngVel_Der}
        \Ddot{\psi} &= \frac{-v+ku}{\tau}
    \end{align}
\end{subequations}

For the cost function, we did not weight any of the inputs as high value, so $P$ and $Q$ were set to identity matrices. The planner's time horizon was 2 seconds with 4 control steps within that time horizon. The spherical harmonics were estimated to the fourth order using approximately 1000 points spread out uniformly over a spherical surface for the least-squares estimator.

The simulation was performed in MATLAB 2020a on a laptop with Intel i7 processor and 32GB of RAM. The optimization of Equation (\ref{LS_Opt}) is done by CasADi. The constrained least squares estimator for Equation (\ref{LS_Weights}) was solved using MATLAB's built-in least squares solver: lsqlin().
\subsection{Results}
\begin{figure}[htbp]
\centerline{\includegraphics[scale=.6]{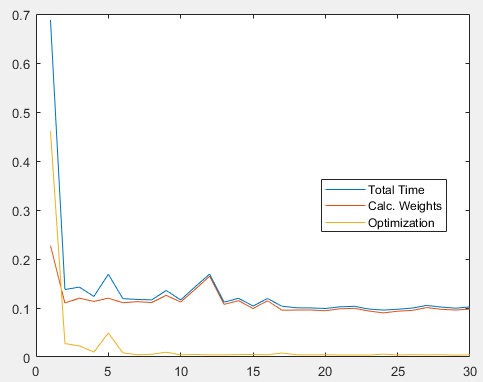}}
\caption{ Timed trial of planning process. The time to calculate the spherical harmonics is in red, to perform the optimization is in yellow, and the total time to create a trajectory is in blue.}
\label{Time_To_Complete}
\end{figure}

The result of our trajectory planner can be seen in Fig. \ref{fig:Path_Over_Time} where the agent reaches its goal by maneuvering through a narrow gap between two obstacles. 
This is in contrast to previous methods, such as \cite{castillo2020real}, that would not reach the goal. 
The planner was able to calculate each time step within 0.2 seconds except for the first step as seen in Fig. \ref{Time_To_Complete}. 
The first step takes much longer because the initial guess was not fine tuned. 
However, it shows once the agent gains knowledge about its surroundings, subsequent planning steps do not take as long.

In the simulation, the agent performs straight-line paths to the goal even if an obstacle is in the way. Rather than going around the obstacle in an arc-like path, the agent gets as close to the obstacle as possible and then slides along the boundaries. 
This action could potentially be ameliorated by adding a repulsive force in the cost function to avoid head on collisions to the obstacles. 
Another solution would be adding a global planner to the system to prevent unnecessary closeness to obstacles. 
It should also be noted that the planner has a smaller time horizon than other planners.
The comparatively small time horizon is due to the SH collision-free space estimation not being able to generate a large volume while being close to an obstacle.
Unlike, other optimization techniques mentioned in this paper, our approach can only plan as far as the agent can see.
However, this is closer to a real world situation where an environment is not known to the agent ahead of time. 
This would result in the agent not being able to see around walls to get to the goal like in \cite{castillo2020real}.

\section{Conclusion}\label{s:conclusion}
We have proposed an optimization-based strategy for collision avoidance (local motion planning) using spherical harmonics to generate a collision-free space around the agent. Our proposed solution approach to the collision avoidance problem was shown to be successful in planning trajectories in real-time where static obstacles are present. 
The efficacy of our approach was demonstrated with an agent trying to navigate through a narrow space where other similar optimization-based planners would fail. 
Due to the limitation of only considering line-of-sight trajectories, the planner cannot have a long time-horizon which results in sub-optimal trajectories where the agent slows down when the goal is being directly blocked by an obstacle. 
However, the agent will eventually get around an obstacle if there is an opening. 
Another benefit of our approach is the number of obstacles has a minuscule effect on the performance of the planner since the planner is only concerned with modeling its free space rather than each individual obstacle.

The next step for improving the trajectory planner is to incorporate dynamic obstacles into the planning framework. 
This will allow our approach to be applied to real-world applications where an environment is dynamic like a warehouse, pedestrian walkway, or outdoors. 
We also intend on applying our approach to physical robots like drones and mobile robots. 
Finally, we plan on testing more advances schemes for estimating the spherical harmonic weights such as an Unscented Kalman Filter.

\bibliography{ifacconf}             % bib file to produce the bibliography
                                                     % with bibtex (preferred)
                                                   
%\begin{thebibliography}{xx}  % you can also add the bibliography by hand

%\bibitem[Able(1956)]{Abl:56}
%B.C. Able.
%\newblock Nucleic acid content of microscope.
%\newblock \emph{Nature}, 135:\penalty0 7--9, 1956.

%\bibitem[Able et~al.(1954)Able, Tagg, and Rush]{AbTaRu:54}
%B.C. Able, R.A. Tagg, and M.~Rush.
%\newblock Enzyme-catalyzed cellular transanimations.
%\newblock In A.F. Round, editor, \emph{Advances in Enzymology}, volume~2, pages
%  125--247. Academic Press, New York, 3rd edition, 1954.

%\bibitem[Keohane(1958)]{Keo:58}
%\newblock \emph{Power and Interdependence: World Politics in Transitions}.
%\newblock Little, Brown \& Co., Boston, 1958.

%\bibitem[Powers(1985)]{Pow:85}
%\newblock Is there a way out?
%\newblock \emph{Harpers}, pages 35--47, June 1985.

%\bibitem[Soukhanov(1992)]{Heritage:92}
%\newblock \emph{{The American Heritage. Dictionary of the American Language}}.
%\newblock Houghton Mifflin Company, 1992.

%\end{thebibliography}

\appendix

\end{document}